\documentclass[lettersize,journal]{IEEEtran}

\usepackage{amsmath,amsfonts,amssymb}

\usepackage{array}
\usepackage{booktabs}
\usepackage{multirow}
\usepackage{multicol}
\usepackage{makecell}

\usepackage{graphicx}
\usepackage{subcaption}
\usepackage[dvipsnames]{xcolor}
\usepackage{tikz}
\usetikzlibrary{arrows.meta,positioning,calc,decorations.pathreplacing}

\usepackage{textcomp}
\usepackage{url}
\usepackage{hyperref}
\usepackage{cite}
\usepackage{stfloats}

\begin{document}

\title{Language-Statistical Analysis of Neural Audio Codec Tokens \\Across Architectures, Corpora, and Noise Conditions}

\author{Joonyong~Park,~\IEEEmembership{Student Member,~IEEE,}
        Shinnosuke~Takamichi,~\IEEEmembership{Member,~IEEE,}
        David~M.~Chan,~\IEEEmembership{Member,~IEEE,}
        Shunsuke~Kando,~\IEEEmembership{Student Member,~IEEE,}
        Yuki~Saito,~\IEEEmembership{Member,~IEEE,}
        and~Hiroshi~Saruwatari,~\IEEEmembership{Member,~IEEE}%
\thanks{J.~Park, S.~Kando, Y.~Saito, and H.~Saruwatari are with the Grad. School of Information Science and Technology, The University of Tokyo, Tokyo, Japan.}%
\thanks{S.~Takamichi is with the Faculty of Science and Technology, Keio University, Tokyo, Japan.}%
\thanks{D.~M.~Chan is with the Berkeley Artificial Intelligence Research Lab (BAIR), University of California, Berkeley, CA, USA.}}%

\markboth{}%
{Park \MakeLowercase{\textit{et al.}}: Language-Statistical Analysis of Neural Audio Codec Tokens Across Architectures, Corpora, and Noise Conditions}

\maketitle

\begin{abstract}
Neural audio codecs (NACs) convert speech into discrete token sequences, and prior work has reported that these sequences follow language-like statistical laws.
This paper analyzes the token statistics of 13 NACs spanning multi-codebook residual vector quantization (RVQ), single-codebook VQ, and non-VQ designs, evaluated on three corpora under clean, white-noise, and real-world DEMAND-noise conditions.
Zipf and Heaps parameters, unigram entropy, codebook occupancy, and Jensen--Shannon divergence (JSD) are estimated from matched token samples with explicit fit-validity safeguards and family-conditional $n$-gram orders.
Corpus identity explains little variance in any metric, whereas acoustic condition and quantizer meta-category dominate in a metric-dependent way, and unigram entropy is the metric most strongly associated with meta-category.
Clean-to-noise JSD computed at a common unigram order is associated with mel-cepstral distortion most clearly under DEMAND noise.
The collapse and explosion degradation signatures previously reported for RVQ codecs concentrate in RVQ cells under white and DEMAND noise, respectively; explosion also occurs in non-VQ codecs, and single-codebook VQ codecs shift in occupancy and distribution shape without either signature.
These results provide architecture-conditioned conventions for applying language-statistical analysis to NAC tokens.
\end{abstract}


\section{Introduction}
\label{sec:introduction}

\IEEEPARstart{N}{eural} audio codecs (NACs) have emerged as a foundational component in modern speech processing pipelines.
By converting continuous acoustic waveforms into sequences of discrete tokens through learned codebooks, NACs enable the direct application of powerful sequence-modeling techniques---originally developed for natural language processing---to speech data~\cite{soundstream, encodec}.
This paradigm has been adopted across a wide range of downstream tasks, including text-to-speech (TTS) synthesis~\cite{speechtokenizer, valle}, automatic speech recognition (ASR)~\cite{codec_asr}, and audio generation~\cite{audiolm}.

Despite their growing importance, a fundamental question remains underexplored: \emph{do NAC token sequences exhibit statistical regularities similar to those found in natural language, and if so, what do these regularities reveal about codec quality and robustness?}
Natural language text is known to follow well-established statistical laws, notably Zipf's law~\cite{zipf1949} governing rank--frequency distributions and Heaps' law~\cite{heaps1978} governing vocabulary growth.
If NAC tokens share these properties, analytical tools developed for natural language become applicable to speech tokens on a known statistical footing: these laws provide compact descriptions of frequency concentration, vocabulary growth, and per-token information content.
If the distributions diverge substantially, conclusions and design heuristics imported from natural language processing may not hold for speech tokens.

Our previous work~\cite{park_asru2025} provided initial evidence that NAC tokens, particularly at the 3-gram level, exhibit Zipfian distributions.
The degree of adherence was further shown to correlate with the resynthesis quality of the same codec, measured by ASR word error rate and by the pseudo-MOS predictor UTMOS~\cite{utmos}.
However, that analysis was limited to clean speech from a single corpus and a small set of NACs based on multi-codebook residual vector quantization (RVQ).
It left open questions about estimation reliability, noise robustness, cross-corpus generalization, and the mechanisms underlying distributional changes.

In this paper, we substantially broaden this line of inquiry along both empirical and methodological dimensions.
On the empirical side, we expand from six RVQ-based NACs to \emph{13 NACs} spanning three quantizer meta-categories: multi-codebook RVQ, single-codebook VQ, and non-VQ alternatives including scalar-quantized, focal-modulation, and semantic designs.
The evaluation covers single-speaker, multi-speaker, and multilingual corpora under clean, white-noise, and DEMAND-noise conditions~\cite{demand}.
In place of the single fixed corpus size of our previous work~\cite{park_asru2025}, a subdivided chunk design makes finite-sample stability explicit.
On the methodological side, we augment the Zipf / Heaps / entropy triad with \emph{codebook occupancy} and \emph{token perplexity}.
We further introduce \emph{Jensen--Shannon divergence (JSD)}---computed between the $n$-gram frequency distributions that the same codec produces on clean and on noise-corrupted versions of the same corpus---as a nonparametric measure of how far noise displaces the token distribution.
Finally, we decompose the variance of every metric with a \emph{three-way ANOVA} over codec quantizer category, corpus, and noise condition.

Building on this broadened design, our contributions fall into three themes.
First, we show that \emph{which factor shapes the token distribution depends on which statistic is measured, while the choice of corpus matters remarkably little throughout}.
On clean speech, the quantizer architecture separates the codec families: semantically informed RVQ designs sit near the word-frequency power-law exponent of natural language, whereas single-codebook designs produce far steeper frequency distributions.
Once noisy conditions enter, the acoustic condition becomes the main driver of the fitted exponents, and its effect differs across architecture families.
Second, we find that clean-to-noisy JSD is a \emph{decoder-free distribution-level diagnostic} whose association with perceived quality is clearest under real-world DEMAND noise.
Third, we identify \emph{collapse} and \emph{explosion} degradation signatures in token sequences through analysis of codebook occupancy, token repetition rates, and transition entropy.
Collapse is concentrated in RVQ codecs under white noise.
Explosion is most frequent in RVQ codecs under DEMAND noise and otherwise appears almost exclusively for the semantic tokenizer.
Single-codebook VQ codecs deform distributionally under noise without either signature.
We further show that the 3-gram optimum reported in~\cite{park_asru2025} is \emph{family-dependent} rather than universal, motivating per-family analysis conventions for non-RVQ codecs.

The remainder of this paper is organized as follows.
Section~\ref{sec:related} reviews related work.
Section~\ref{sec:methodology} defines the statistical metrics and analysis framework.
Section~\ref{sec:setup} describes the experimental setup.
Section~\ref{sec:reliability} establishes estimation reliability.
Sections~\ref{sec:cross_corpus}--\ref{sec:quality} present cross-architecture, cross-corpus, and quality correlation analyses.
Section~\ref{sec:mechanism} characterizes architecture-dependent degradation signatures.
Section~\ref{sec:discussion} discusses implications and limitations, and Section~\ref{sec:conclusion} concludes this paper.

\section{Related Work}
\label{sec:related}

\subsection{Discrete Tokens from Speech}
\label{ssec:discrete_tokens}

Self-supervised learning (SSL) models such as HuBERT~\cite{hubert} and wav2vec~2.0~\cite{wav2vec2} learn frame-level speech representations from unlabeled audio.
Although these representations are themselves continuous, they are commonly discretized---typically by $k$-means clustering---to obtain token sequences for downstream processing.
The resulting tokens capture phonetic and speaker-related information~\cite{mohamed2022ssl}, making them effective for tasks including spoken language modeling and speech recognition.

NACs offer a complementary approach, producing discrete tokens explicitly optimized for high-fidelity audio reconstruction.
Representative models such as SoundStream~\cite{soundstream} and EnCodec~\cite{encodec} employ encoder--quantizer--decoder architectures with RVQ, compressing audio into multi-level token sequences at various bitrates.
Subsequent work has expanded this design space considerably.
SpeechTokenizer~\cite{speechtokenizer} disentangles semantic and acoustic information across RVQ layers, HiFi-Codec~\cite{hifi_codec} introduces group-RVQ for improved fidelity, and AudioDec~\cite{audiodec} targets streaming scenarios.
FunCodec~\cite{funcodec} provides a flexible open-source toolkit, and DAC~\cite{dac} achieves high compression ratios with improved training on the basis of generative adversarial networks (GANs)~\cite{gan}.
More recent designs explore fundamentally different architectures.
WavTokenizer~\cite{wavtokenizer} and BigCodec~\cite{bigcodec} adopt large single-codebook designs, XCodec2~\cite{xcodec2} uses a massive 65{,}536-entry vocabulary, and SQCodec~\cite{sqcodec} replaces vector quantization with scalar quantization.
Mimi~\cite{mimi} combines RVQ with semantic objectives for real-time dialogue, FocalCodec~\cite{focalcodec} applies focal modulation for low-bitrate coding, and S3Tokenizer~\cite{s3tokenizer} produces semantic-only tokens without a decoder.
A recent survey by Guo et al.~\cite{guo2025recent} provides a comprehensive overview of this rapidly diversifying design space.

The availability of discrete speech tokens has enabled a new generation of language-model-based speech processing systems.
AudioLM~\cite{audiolm} demonstrated that autoregressive modeling of audio tokens can generate coherent speech continuations, while VALL-E~\cite{valle} showed that treating TTS as a language modeling problem over codec tokens enables zero-shot speaker adaptation.
Codec tokens have also been integrated into ASR pipelines~\cite{codec_asr}, where compressed representations reduce computational costs while maintaining recognition accuracy.
This growing reliance on NAC tokens as the interface between speech and language modeling motivates a deeper understanding of their distributional structure.

\subsection{Statistical Laws of Language}
\label{ssec:stat_laws}

Zipf's law~\cite{zipf1949} is among the most robust empirical regularities observed in natural language: the frequency of the $r$-th ranked word in a corpus follows a power law $f(r) \propto r^{-s}$, with $s \approx 1$ at the word level.
Equivalently, the distribution of word frequencies follows a power law with exponent $\alpha = 1 + 1/s \approx 2$~\cite{newman2005power}.
The phenomenon was formalized by Mandelbrot~\cite{mandelbrot1953} and has been confirmed across typologically diverse languages~\cite{gelbukh2001}.
Complementarily, Heaps' law~\cite{heaps1978} describes sublinear vocabulary growth as a function of corpus size, reflecting the diminishing rate at which new word types are introduced.
Shannon entropy~\cite{shannon1951} provides a third lens, quantifying the average information content per symbol and enabling analysis of coding efficiency and redundancy.
Together, these three measures---frequency distribution shape, vocabulary growth rate, and information density---provide a comprehensive characterization of the statistical structure of any symbolic system.

While originally studied in human language, these statistical laws exhibit remarkable universality.
Power-law distributions have been documented in domains ranging from city population sizes and citation counts to internet traffic patterns~\cite{newman2005power, clauset2009powerlaw}.
More recently, Chan et al.~\cite{chan2024visual_tokens} demonstrated that tokens produced by visual codecs (VQ-VAE models for image compression) also exhibit Zipf-like and Heaps-like behavior, and that the degree of adherence to these laws correlates with downstream task performance.
Zipf- and Heaps-like behavior can therefore arise in learned visual token systems as well as in text.

\subsection{Statistical Analysis of Speech Tokens}
\label{ssec:speech_token_analysis}

The statistical analysis of discrete speech tokens has attracted growing attention as these tokens now serve as the direct modeling target of speech language models.
Their frequency structure determines how reliably $n$-gram statistics and neural language models can be estimated from a given amount of speech.
Takamichi et al.~\cite{takamichi2024} conducted the first systematic investigation of whether speech tokens follow Zipf's law, focusing on SSL-based representations from HuBERT.
Their results confirmed power-law behavior in token frequency distributions, providing evidence that discrete speech tokens share fundamental statistical properties with natural language text.
Sicherman and Adi~\cite{sicherman2023} further examined the interpretability and redundancy of SSL-based speech tokens, revealing strong correlations between learned tokens and phonemic categories.
They proposed deduplication and redundancy-reduction techniques that improved downstream performance in spoken language modeling, highlighting the practical implications of understanding token distributional structure.
Most recently, Ashihara et al.~\cite{ashihara2025domain} extended this line of analysis along the domain axis, showing through rank--frequency, perplexity, and TF-IDF analyses that both SSL-based semantic tokens and NAC tokens retain power-law structure across the speech, music, and general-sound domains, while their token-usage patterns remain domain-dependent.

While these studies focused on SSL-based tokens or on cross-domain comparisons, the statistical properties of NAC tokens under controlled within-speech variation have received comparatively less attention.
Our previous study~\cite{park_asru2025} presented the first systematic analysis of NAC token distributions across six NACs in 15 model configurations that vary bitrate and sampling rate, demonstrating that 3-gram token sequences exhibit Zipf and Heaps law behavior similar to natural language.
The degree of Zipfian adherence ($\alpha$ approaching 2 with lower Kolmogorov--Smirnov (KS) distance~\cite{ks_test}) was found to correlate with resynthesis quality metrics such as word error rate (WER) and UTMOS~\cite{utmos}.

However, these prior studies share limitations that motivate the present work.
The analysis in~\cite{park_asru2025} was confined to a single clean corpus, and the condition-level correlation coefficients were modest ($|r| < 0.3$).
Estimation reliability, multi-corpus generalizability, and the mechanisms underlying distributional degradation remained unexplored.

\subsection{Noise Robustness of Neural Audio Codecs}
\label{ssec:noise_robustness}

The robustness of NAC models to acoustic degradation is a critical practical concern, as real-world audio frequently contains environmental noise, reverberation, and channel distortions.
Most NAC models are trained predominantly on clean or studio-quality speech~\cite{guo2025recent}, and their behavior under mismatched conditions---where input audio quality deviates substantially from training data---remains insufficiently characterized.

Existing codec evaluation frameworks such as Codec-SUPERB~\cite{codec_superb} provide standardized benchmarks for comparing codec quality across multiple downstream tasks, but focus predominantly on clean input conditions.
Wu et al.~\cite{codec_overview} surveyed the broader landscape of audio language modeling, noting the importance of codec robustness without providing systematic degradation analysis.
While individual codec papers report reconstruction quality under controlled conditions, no prior work has systematically examined how the \emph{distributional structure} of codec tokens---as characterized by Zipf's law, Heaps' law, and entropy---changes under noise.
This gap matters because distributional metrics can be computed directly from token sequences without audio resynthesis, making them candidate decoder-free, token-space indicators of codec behavior.

\section{Methodology}
\label{sec:methodology}

The present study addresses the gaps identified above through a fully crossed design: 13 codecs spanning three quantizer meta-categories are evaluated on three corpora under a uniform three-condition protocol (clean, white noise, DEMAND noise).
Every token stream then passes through a single pipeline of deduplication, family-conditional $n$-gram mapping, and chunk-based estimation; the sequence-level diagnostics of Section~\ref{sec:mechanism} are computed before deduplication.
On this grid, we quantify estimation reliability explicitly and decompose the variance of every distributional metric across meta-category, corpus, and condition.
We further evaluate JSD as a decoder-free degradation diagnostic alongside resynthesis-quality metrics.
This section defines the pipeline, the metrics, and the analysis framework that implement this design.

\subsection{Token Extraction Pipeline}
\label{ssec:pipeline}

We adopt the Codec-SUPERB framework~\cite{codec_superb} for standardized token extraction across all NACs.
Each NAC processes input audio and produces a token sequence.
For RVQ-based NACs the output has one level per residual codebook, and we analyze the first level, which is empirically validated in Section~\ref{sssec:rvqlevel}.
For single-codebook, scalar-quantized, or semantic tokenizers, we use the single available token stream.

From the extracted tokens, we construct $n$-gram sequences using a family-conditional order chosen on the basis of the minimum-KS ablation of Section~\ref{ssec:ngram_ablation} (column $n$ of Table~\ref{tab:codecs}).
Because only the first stream is analyzed, each utterance yields a one-dimensional token sequence.
Following~\cite{park_asru2025}, we apply \emph{consecutive deduplication} to this sequence: runs of identical tokens are collapsed to a single instance, which removes frame-level repetitions that would bias $n$-gram frequency counts.
The effect of this step is quantified in Section~\ref{sssec:dedup}.
Deduplicated utterance sequences are then concatenated into corpus-level streams for the chunk analysis of Section~\ref{ssec:analysis_framework}.

\subsection{Statistical Metrics}
\label{ssec:metrics}

\subsubsection{Zipf's Law}
Zipf's law states that the frequency $f(r)$ of the $r$-th ranked element in a corpus follows a power law:
\begin{equation}
    f(r) = a \cdot r^{-s},
    \label{eq:zipf}
\end{equation}
where $a > 0$ is a scaling constant and $s > 0$ is the rank--frequency exponent; natural language text typically exhibits $s \approx 1$ at the word level~\cite{zipf1949}.
We quantify this behavior through the equivalent frequency-distribution form: under Eq.~\ref{eq:zipf}, the occurrence counts $x$ of the individual $n$-gram types follow $p(x) \propto x^{-\alpha}$ with $\alpha = 1 + 1/s$~\cite{newman2005power}, so the word-level reference value corresponds to $\alpha \approx 2$.
Every Zipf exponent reported in this paper is this frequency-distribution exponent $\alpha$.

We estimate $\alpha$ using maximum likelihood estimation (MLE) over the multiset of type frequencies via the \texttt{powerlaw} library~\cite{powerlaw_lib}:
\begin{equation}
    \hat{\alpha}_{\text{MLE}} = 1 + \frac{n}{\sum_{i=1}^{n} \log(x_i / x_{\min})},
    \label{eq:alpha_mle}
\end{equation}
where $x_i$ is the occurrence count of the $i$-th $n$-gram type, $x_{\min}$ is the minimum-frequency threshold selected by the library, and $n$ is the number of types with $x_i \geq x_{\min}$.
Goodness of fit is assessed using the KS distance between the empirical and fitted cumulative distribution functions of the type frequencies; lower KS values indicate better fit.
Two safeguards apply to every fit in this paper.
First, whenever the library's default search bound truncates the estimate at $\alpha = 3$, we refit with the parameter range expanded to $\alpha \in [1.01, 8]$.
Second, we treat a fit as \emph{undefined} when the type--token ratio $V/N$ exceeds $0.9$.
In this regime, the observed $n$-gram inventory is nearly unique relative to the sample size and the frequency profile is degenerate, so the estimator produces arbitrarily steep exponents with spuriously small KS values.

\subsubsection{Heaps' Law}
Heaps' law~\cite{heaps1978} describes the sublinear growth of vocabulary size $V$ as a function of the total number of tokens $N$:
\begin{equation}
    V(N) = K \cdot N^{\beta}, \quad 0 < \beta < 1,
    \label{eq:heaps}
\end{equation}

where $K> 0$ is a scaling factor reflecting the initial vocabulary introduction rate, and $\beta$ determines how quickly new types continue to appear.
For word-level English text, $K$ typically lies in the range $10$--$100$~\cite{heaps1978}.
Values of $\beta$ close to 1 indicate near-linear vocabulary growth (high diversity), while lower values suggest vocabulary saturation.
Word-level natural language text typically exhibits $\beta \approx 0.4$--$0.6$~\cite{heaps1978}.

We estimate $\beta$ and $K$ via ordinary least squares (OLS) regression on log-transformed data:
\begin{equation}
    \log V(N) = \log K + \beta \cdot \log N.
    \label{eq:heaps_ols}
\end{equation}

\subsubsection{Shannon Entropy and Redundancy}
The Shannon entropy of a token sequence measures the average information content per token:
\begin{equation}
    H = -\sum_{i=1}^{|V|} p_i \log_2 p_i,
    \label{eq:entropy}
\end{equation}
where $p_i$ is the probability of the $i$-th token type and $|V|$ is the number of observed token types.
The redundancy $R$ quantifies how far the observed entropy falls from the theoretical maximum under a uniform distribution:
\begin{equation}
    R = \frac{\log_2 |V| - H}{\log_2 |V|},
    \label{eq:redundancy}
\end{equation}
where $\log_2 |V|$ is the maximum entropy achievable over the observed vocabulary.
For printed English, Shannon estimated a redundancy of roughly one half, corresponding to about one bit of information per letter~\cite{shannon1951}.
We also report the \emph{token perplexity} $2^H$, which re-expresses the entropy as the effective number of frequently used token types under the unigram distribution.

\subsubsection{JSD}
To quantify how noise alters the token distribution relative to clean speech, we compute JSD.
JSD serves as a nonparametric measure computed over the empirical frequency distribution of \emph{all observed $n$-gram types}, rather than over a small number of fitted summary parameters such as $\alpha$ or $\beta$.
Given a clean reference distribution $P$ and a degraded distribution $Q$ over the shared $n$-gram vocabulary, JSD is defined as:
\begin{equation}
    \text{JSD}(P \| Q) = \tfrac{1}{2} D_{\text{KL}}(P \| M) + \tfrac{1}{2} D_{\text{KL}}(Q \| M),
    \label{eq:jsd}
\end{equation}
where $M = \tfrac{1}{2}(P + Q)$ is the mixture distribution and $D_{\text{KL}}$ denotes the Kullback--Leibler divergence.
JSD is bounded in $[0, 1]$, symmetric, and well-defined even when $P$ and $Q$ have non-overlapping support.

\subsection{Quality Metrics} 
\label{ssec:quality_metrics}

Each codec is driven at the sampling rate it was trained for.
Thus, the corpus waveform is resampled to that rate before encoding.
Resynthesis uses each codec's complete standard decoding path: the original, non-deduplicated token sequences at all codebook levels are decoded to a waveform.
The decoded waveform is then resampled back to 16\,kHz, at which every metric below is computed.
We evaluate resynthesized speech using five metrics:
\begin{itemize}
    \item \textbf{dCER}: Differential character error rate. Whisper-large-v3~\cite{whisper} transcribes both the clean ground-truth waveform and the resynthesized waveform. The character error rate between the two transcripts is computed per utterance after multilingual text normalization and capped at 1.0.
    \item \textbf{UER}: Unit error rate, the Levenshtein distance between the token sequences obtained from the clean and noise-corrupted versions of the same utterance, normalized by the clean sequence length. UER is defined only under noisy conditions.
    \item \textbf{UTMOS}: Pseudo-MOS score estimating perceived naturalness~\cite{utmos}.
    \item \textbf{MCD}: Standard mel-cepstral distortion in dB, computed from mel-cepstral features extracted with WORLD~\cite{world}, with DTW alignment, using the clean ground-truth waveform as reference. The distance includes the zeroth mel-cepstral coefficient, so the resynthesis is rescaled to the root-mean-square level of its reference.
    \item \textbf{F0 error}: F0 is extracted with Praat's autocorrelation method~\cite{praat} (search range 50--1,100\,Hz, voicing threshold 0.6). The two contours are aligned by DTW as for MCD, and the error is taken over frames voiced in both signals. We report the per-utterance median absolute error (F0MedAE) alongside the conventional root-mean-square error.
\end{itemize}

\subsection{Analysis Framework}
\label{ssec:analysis_framework}

\subsubsection{Chunk Analysis}
\label{sssec:chunk}
We partition each corpus-level token stream into non-overlapping chunks of increasing size (5K to 500K tokens) and compute the Zipf/Heaps parameters once per chunk; deviations from the matched 500K-token reference of Section~\ref{ssec:convergence} quantify estimation reliability, i.e., finite-sample stability.

\subsubsection{Codebook Occupancy}
As an additional diagnostic capturing token-usage concentration rather than distribution shape, we report \emph{codebook occupancy}: the number of unique tokens that appear in a sequence, normalized by an estimate of the codebook size.
Because the nominal codebook size is unavailable for some non-VQ codecs (column $|V|$ of Table~\ref{tab:codecs}), the denominator is the largest observed token index plus one for every codec.
This proxy assumes contiguous token indexing; for codecs without a nominal codebook size, we interpret occupancy only through within-codec changes across conditions.

\subsubsection{Variance Decomposition}
We perform a \emph{three-way ANOVA} with codec quantizer category, corpus, and noise condition as factors.
The ANOVA decomposes the variance in the distributional parameters, with effect sizes reported via $\eta^2$.
Each codec contributes multiple cells across corpora and conditions, so the reported $\eta^2$ values are descriptive variance partitions over the cell grid rather than tests on independent samples.
A linear mixed model with a codec random intercept serves as the corresponding robustness check.

\section{Experimental Setup}
\label{sec:setup}

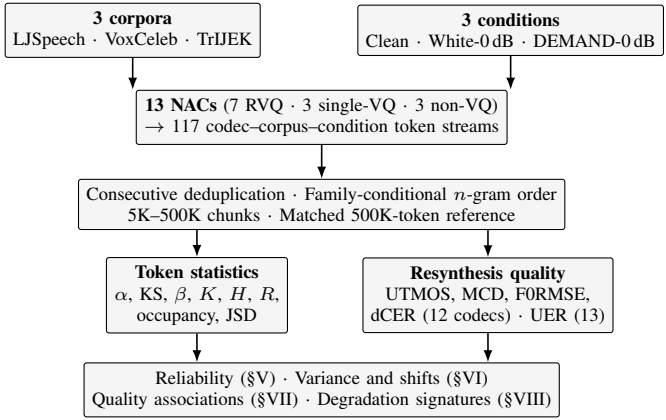
\begin{figure}[t]
\centering
\begin{tikzpicture}[
  font=\scriptsize,
  box/.style={draw, rounded corners=1pt, align=center, inner sep=3pt, fill=gray!8},
  head/.style={font=\scriptsize\bfseries},
  arr/.style={-{Latex[length=1.6mm]}, semithick}
]
\node[box] (corpora) at (0,0) {\textbf{3 corpora}\\ LJSpeech $\cdot$ VoxCeleb $\cdot$ TrIJEK};
\node[box] (cond) at (5.0,0) {\textbf{3 conditions}\\ Clean $\cdot$ White-0\,dB $\cdot$ DEMAND-0\,dB};
\node[box] (nacs) at (2.5,-1.15) {\textbf{13 NACs} (7 RVQ $\cdot$ 3 single-VQ $\cdot$ 3 non-VQ)\\ $\rightarrow$ 117 codec--corpus--condition token streams};
\node[box] (proc) at (2.5,-2.3) {Consecutive deduplication $\cdot$ Family-conditional $n$-gram order\\ 5K--500K chunks $\cdot$ Matched 500K-token reference};
\node[box] (dist) at (0.85,-3.5) {\textbf{Token statistics}\\ $\alpha$, KS, $\beta$, $K$, $H$, $R$,\\ occupancy, JSD};
\node[box] (qual) at (4.7,-3.5) {\textbf{Resynthesis quality}\\ UTMOS, MCD, F0RMSE,\\ dCER (12 codecs) $\cdot$ UER (13)};
\node[box] (ana) at (2.5,-4.75) {Reliability (\S\ref{sec:reliability}) $\cdot$ Variance and shifts (\S\ref{sec:cross_corpus})\\ Quality associations (\S\ref{sec:quality}) $\cdot$ Degradation signatures (\S\ref{sec:mechanism})};
\draw[arr] (corpora.south) -- (nacs.north -| corpora.south);
\draw[arr] (cond.south) -- (nacs.north -| cond.south);
\draw[arr] (nacs) -- (proc);
\draw[arr] (proc.south -| dist.north) -- (dist.north);
\draw[arr] (proc.south -| qual.north) -- (qual.north);
\draw[arr] (dist.south) -- (ana.north -| dist.south);
\draw[arr] (qual.south) -- (ana.north -| qual.south);
\end{tikzpicture}
\caption{Overview of the experimental design.}
\label{fig:overview}
\end{figure}

Figure~\ref{fig:overview} summarizes the evaluation design and the analysis flow of Sections~\ref{sec:reliability}--\ref{sec:mechanism}.

\subsection{NACs}
\label{ssec:codecs}

We evaluate 13 NAC models spanning three quantizer meta-categories, as summarized in Table~\ref{tab:codecs}.
All 13 codecs are analyzed on the same three corpora $\times$ three conditions grid described in Sections~\ref{ssec:corpora}--\ref{ssec:noise}, which yields a total of $117$ codec--corpus--condition cells.
Each codec is evaluated in the default configuration of its released implementation, without manual bandwidth adjustment (column \textbf{kbps} of Table~\ref{tab:codecs}).

For analytical purposes, we group the 13 codecs into three \emph{meta-categories} based on quantizer topology (column~\textbf{M} of Table~\ref{tab:codecs}):
(R) \emph{multi-codebook RVQ} --- classic 1{,}024-entry RVQ designs (SpeechTokenizer, AcademiCodec, AudioDec, EnCodec, FunCodec, DAC-24k) plus the hybrid RVQ codec Mimi~\cite{mimi};
(S) \emph{single-codebook VQ} --- designs using a single but larger codebook (BigCodec~\cite{bigcodec}, WavTokenizer~\cite{wavtokenizer}, XCodec2~\cite{xcodec2});
and (N) \emph{non-VQ alternatives} --- architectures that replace or bypass standard vector quantization (SQCodec~\cite{sqcodec}: scalar quantization; FocalCodec~\cite{focalcodec}: focal modulation; S3Tokenizer~\cite{s3tokenizer}: semantic-only).
This grouping enables family-aware analysis of how codebook size, quantizer type, and architectural choice affect distributional properties.

\begin{table}[t]
\caption{NAC configurations. \textbf{Family}: architectural family. \textbf{M}: meta-category (\textbf{R}: multi-codebook RVQ; \textbf{S}: single-codebook VQ; \textbf{N}: non-VQ). \textbf{$|V|$}: codebook vocabulary size (`var.': no nominal size; the observed-token proxy of Section~\ref{ssec:analysis_framework} is used). \textbf{$n$}: $n$-gram order used in all analyses. \textbf{SR}: sampling rate. \textbf{kbps}: bitrate of the evaluated configuration (`--': not applicable or not published). $\dagger$: semantic-only, no decoder; audio-based quality metrics unavailable.}
\label{tab:codecs}
\centering
\scriptsize
\begin{tabular}{@{}llcrccc@{}}
\toprule
\textbf{Codec} & \textbf{Family} & \textbf{M} & $|V|$ & $n$ & \textbf{SR} & \textbf{kbps} \\
\midrule
SpeechTokenizer & Classic RVQ & R & 1{,}024 & 3 & 16k & 4 \\
AcademiCodec & Classic RVQ & R & 1{,}024 & 3 & 24k & 3 \\
AudioDec & Classic RVQ & R & 1{,}024 & 3 & 24k & 6.4 \\
EnCodec & Classic RVQ & R & 1{,}024 & 3 & 24k & 6 \\
FunCodec & Classic RVQ & R & 1{,}024 & 3 & 16k & 16 \\
DAC-24k & Classic RVQ & R & 1{,}024 & 3 & 24k & 24 \\
Mimi & Hybrid RVQ & R & 2{,}048 & 2 & 24k & 4.4 \\
BigCodec & Large single & S & 8{,}192 & 2 & 16k & 1.04 \\
WavTokenizer & Large single & S & 4{,}096 & 2 & 24k & -- \\
XCodec2 & Massive vocab & S & 65{,}536 & 1 & 16k & -- \\
SQCodec & Scalar quant. & N & var. & 1 & 16k & 3 \\
FocalCodec & Focal mod. & N & 8{,}192 & 1 & 16k & 0.65 \\
S3Tokenizer$^\dagger$ & Semantic & N & 4{,}096 & 3 & 16k & -- \\
\bottomrule
\end{tabular}
\end{table}

\subsection{Speech Corpora}
\label{ssec:corpora}

We use three corpora selected to cover three practically relevant sources of variation beyond acoustic condition: speaker diversity, language diversity, and recording style (Table~\ref{tab:corpora}). All corpora are approximately matched at $\sim$10 hours of speech. 
LJSpeech~\cite{ljspeech} provides a single-speaker monolingual reference, and VoxCeleb~\cite{voxceleb} adds multi-speaker variability.
TrIJEK contributes multilingual coverage (Japanese, Korean, English) recorded by a single speaker, which reduces speaker variation across the three languages.
All 13 codecs are evaluated on the full three-corpora set.

\begin{table}[t]
\caption{Speech corpora used in this study.}
\label{tab:corpora}
\centering
\scriptsize
\begin{tabular}{@{}llccc@{}}
\toprule
\textbf{Corpus} & \textbf{Category} & \textbf{Language} & \textbf{Speakers} & \textbf{Hours} \\
\midrule
LJSpeech & Single-speaker mono & English & 1        & $\sim$10 \\
VoxCeleb & Multi-speaker       & English & Many     & $\sim$10 \\
TrIJEK   & Multilingual mono   & JA/KO/EN & 1       & $\sim$10 \\
\bottomrule
\end{tabular}
\end{table}

\subsection{Noise Conditions}
\label{ssec:noise}

To assess noise robustness, we evaluate two noisy variants of each corpus alongside the clean condition:

\begin{itemize}
    \item \textbf{Clean}: the original recordings with no additional noise.
    \item \textbf{White Noise}: Gaussian white noise added at 0\,dB SNR.
    \item \textbf{DEMAND Noise}: real-world noise from the DEMAND corpus~\cite{demand} added at 0\,dB SNR. For each utterance, one of five DEMAND environments (office meeting, cafeteria, restaurant, bus, metro) is drawn at random, together with a random recording channel and start offset. Each cell therefore pools several realistic environments.
\end{itemize}

Noisy waveforms are generated once per corpus with a fixed random seed, and the identical noisy waveform is presented to every codec.
Cross-codec differences under noise therefore do not reflect noise-realization variability.
Clean is treated as the reference condition for computing all clean-to-noisy deviation metrics.

\section{Estimation Reliability and Analysis Configuration}
\label{sec:reliability}

\subsection{Finite-Sample Stability}
\label{ssec:convergence}

Before comparing codecs and acoustic conditions, we assess the finite-sample stability of the distributional parameters at the sample scale used in the subsequent experiments, using the chunk analysis defined in Section~\ref{sssec:chunk}.
For each codec--corpus--condition cell, we calculate each parameter from non-overlapping chunks of increasing size and compare it with the estimate obtained from a matched 500K-token sample.
For the Heaps intercept $K$, whose scale varies substantially across codecs, we use the absolute log-ratio rather than the raw absolute difference.
A matched reference is available for 114 of the 117 cells; the three Mimi--TrIJEK cells are excluded because their token streams do not reach this size.

\begin{figure*}[t]
    \centering
    \includegraphics[width=0.98\textwidth]{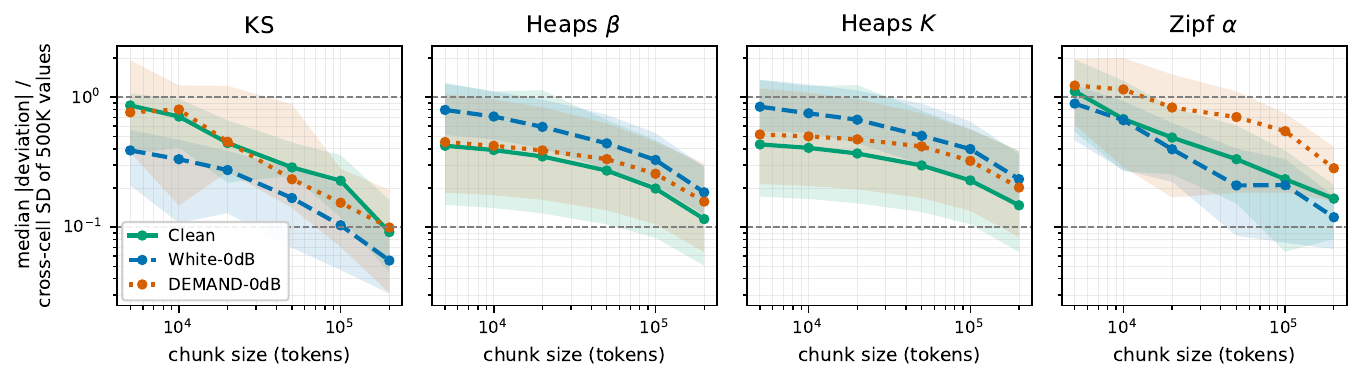}
    \caption{Finite-sample stability across the 114 codec--corpus--condition cells with a matched 500K-token reference. Lines and shaded regions show the median and interquartile range of the absolute deviation from that reference (absolute log-ratio for $K$), normalized by the cross-cell standard deviation of the reference values so that the four panels share one scale. Dashed guides mark deviations of 0.1 and 1.0 times that spread.}
    \label{fig:convergence}
\end{figure*}

Figure~\ref{fig:convergence} summarizes the 114 cells with a matched reference.
As the four parameters have different units and scales, each deviation is divided by the cross-cell standard deviation of the 500K-token reference values, so that all panels share one scale.
The KS distance converges fastest.
Pooled across conditions, its median deviation falls from 0.55 of the cross-cell spread at 5K tokens to 0.07 at 200K, making it the only parameter to reach the lower 0.1 guide.
$\beta$ and $\alpha$ remain at 0.17 and $K$ at 0.19 at 200K tokens.
At the opposite end, $\alpha$ is the least stable parameter at 5K tokens, where its median deviation 1.06 exceeds the full cross-cell spread.
We therefore use matched 500K-token samples as the common reference scale for all downstream comparisons.

\subsection{Selection and Sensitivity of $n$-gram Order}
\label{ssec:ngram_ablation}

\begin{figure}[t]
    \centering
    \includegraphics[width=0.98\linewidth]{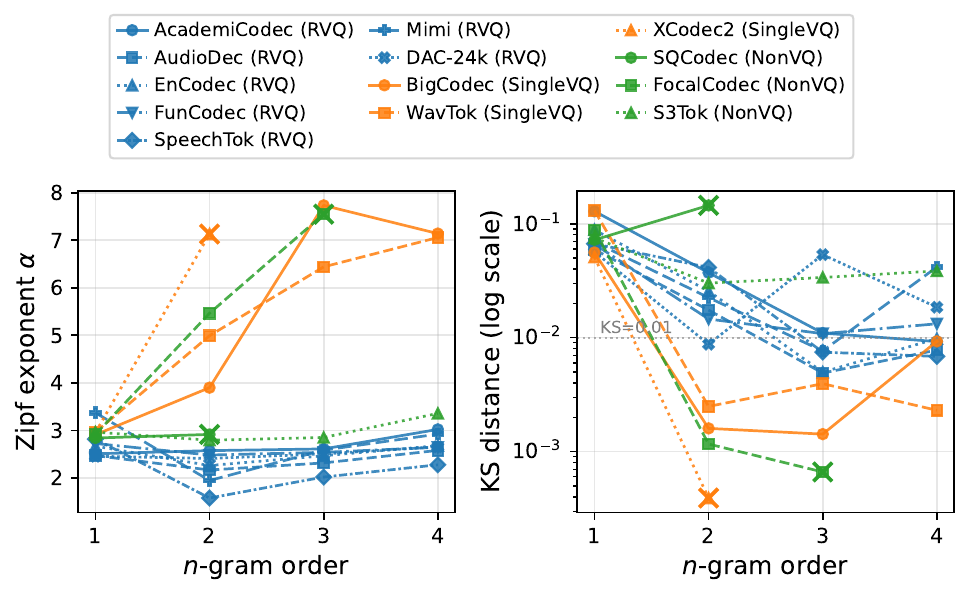}
    \caption{Effect of $n$-gram order on the Zipf exponent $\alpha$ and KS fit quality, averaged over three clean corpora. For codecs whose higher orders fail the $V/N$ validity criterion the sweep terminates early; the last valid order of such series is marked with $\times$.}
    \label{fig:ngram_sweep}
\end{figure}

Sweeping the $n$-gram order for all 13 codecs, with the fitting safeguards of Section~\ref{ssec:metrics}, shows that the optimal order is architecture-dependent, as illustrated in Fig.~\ref{fig:ngram_sweep}.
Most multi-codebook RVQ codecs favor $n=3$ or $4$, with DAC-24k as the exception (Table~\ref{tab:ngram_percorpus}).
Single-codebook and large-vocabulary codecs favor lower orders because their higher-order inventories rapidly approach $V \approx N$, beyond which the fit is undefined.

\begin{table}[t]
\caption{Per-corpus $n$-gram order selection. \textbf{M}: meta-category, where R, S, and N denote multi-codebook RVQ, single-codebook VQ, and non-VQ codecs. The remaining columns give the KS-optimal order for each clean corpus, the order minimizing the corpus-mean KS, and the order used in the main analyses, computed under the fitting safeguards of Section~\ref{ssec:metrics}. $^*$: the adopted order deviates from the corpus-mean optimum by one; $^\dagger$: by two.}
\label{tab:ngram_percorpus}
\centering
\scriptsize
\begin{tabular}{@{}llccccc@{}}
\toprule
\textbf{Codec} & \textbf{M} & \textbf{LJS} & \textbf{TrIJEK} & \textbf{Vox} & \textbf{Argmin $\overline{\text{KS}}$} & \textbf{Used} \\
\midrule
AcademiCodec$^*$      & R & 4 & 3 & 4 & 4 & 3 \\
AudioDec              & R & 3 & 3 & 3 & 3 & 3 \\
DAC-24k$^*$           & R & 2 & 2 & 2 & 2 & 3 \\
EnCodec               & R & 3 & 3 & 3 & 3 & 3 \\
FunCodec              & R & 3 & 3 & 2 & 3 & 3 \\
Mimi$^\dagger$        & R & 4 & 3 & 3 & 4 & 2 \\
SpeechTokenizer$^*$   & R & 4 & 4 & 4 & 4 & 3 \\
BigCodec$^*$          & S & 1 & 1 & 1 & 1 & 2 \\
WavTokenizer          & S & 1 & 2 & 1 & 2 & 2 \\
XCodec2               & S & 1 & 1 & 1 & 1 & 1 \\
FocalCodec            & N & 1 & 1 & 1 & 1 & 1 \\
S3Tokenizer           & N & 3 & 3 & 3 & 3 & 3 \\
SQCodec               & N & 1 & 1 & 1 & 1 & 1 \\
\bottomrule
\end{tabular}
\end{table}

The analysis orders used in the main experiments were fixed before this ablation and are retained for consistency with the completed analyses.
Table~\ref{tab:ngram_percorpus} reports the per-corpus optima: the adopted order matches the corpus-mean KS optimum for eight of the 13 codecs, deviates by one order for four codecs, and by two orders for Mimi.
Table~\ref{tab:fixed_n} quantifies the cross-family $\alpha$ comparison at fixed common orders.
At the unigram order, which covers the full grid, single-codebook VQ codecs have a higher mean $\alpha$ than multi-codebook RVQ codecs, but the meta-category effect is not significant.
At $n=2$ the separation is large and significant, but valid fits remain for only 25 of the 39 codec--corpus cells, and the single-codebook VQ family retains a single valid cell.
We therefore use the adopted orders for within-codec analyses and treat absolute cross-family differences under codec-specific orders as descriptive.

\begin{table}[t]
\caption{Fixed-order sensitivity of the cross-family $\alpha$ comparison on clean speech: number of codec--corpus cells with valid fits, per-family mean$\pm$std $\alpha$, and the one-way meta-category effect size at each common order. $^\ddag$: a single valid cell remains, so no std is defined.}
\label{tab:fixed_n}
\centering
\scriptsize
\begin{tabular}{@{}cccccc@{}}
\toprule
\textbf{Order} & $N$ & $\overline{\alpha}_{\mathrm{RVQ}}$ & $\overline{\alpha}_{\mathrm{SingleVQ}}$ & $\overline{\alpha}_{\mathrm{NonVQ}}$ & $\eta^2_{\text{meta}}$ ($p$) \\
\midrule
$n=1$ & 39 & 2.71$\pm$0.63 & 2.97$\pm$0.03 & 2.88$\pm$0.11 & 0.05 ($0.37$) \\
$n=2$ & 25 & 2.24$\pm$0.41 & 4.39$^\ddag$ & 2.99$\pm$0.01 & 0.63 ($<$$0.001$) \\
\bottomrule
\end{tabular}
\end{table}

\subsection{Pipeline Validation}
\label{ssec:pipeline_val}

Beyond the $n$-gram order, the extraction pipeline of Section~\ref{ssec:pipeline} involves two further choices: the analyzed RVQ level and consecutive deduplication. We validate those in Table~\ref{tab:pipeline_val}.

\begin{table}[t]
\caption{Pipeline validation on the five multi-codebook RVQ codecs with level-wise extractions (first-level 3-gram streams; 105 codec--condition cells). (a) KS fit distance by RVQ level (mean$\pm$std). (b) Effect of consecutive deduplication on the first-level stream; $p$ from paired $t$-tests.}
\label{tab:pipeline_val}
\centering
\scriptsize
\begin{tabular}{@{}lcccc@{}}
\toprule
\multicolumn{5}{@{}l}{\textit{(a) KS distance by RVQ level}}\\
\textbf{Level} & 1st & 2nd--4th & 5th--8th & 9th+ \\
\midrule
KS & 0.012$\pm$0.007 & 0.049$\pm$0.068 & 0.110$\pm$0.095 & 0.105$\pm$0.052 \\
\midrule
\multicolumn{5}{@{}l}{\textit{(b) Deduplication effect on the first level}}\\
 & $\alpha$ & KS & $\beta$ & $K$ \\
\midrule
Without dedup & 2.26 & 0.012 & 0.81 & 4.72 \\
With dedup & 2.33 & 0.012 & 0.82 & 4.63 \\
Cohen's $d$ ($p$) & 0.46 ($<$$0.001$) & $-$0.06 ($0.54$) & 0.49 ($<$$0.001$) & $-$0.05 ($0.58$) \\
\bottomrule
\end{tabular}
\end{table}

\subsubsection{RVQ-Level Selection}
\label{sssec:rvqlevel}
For multi-codebook RVQ codecs, we use the first residual-quantizer level.
In the five codecs with level-wise results, the first level has a lower KS distance than later levels (Table~\ref{tab:pipeline_val}a). Thus, we retain this convention for the full codec set.

\subsubsection{Deduplication}
\label{sssec:dedup}
We apply consecutive deduplication before fitting the distributional models, following~\cite{park_asru2025}.
On the same cells, deduplication increases the estimated $\alpha$ and $\beta$ with a medium standardized effect (Cohen's $d \approx 0.5$) but does not significantly change KS or $K$ (Table~\ref{tab:pipeline_val}b).



\section{Cross-Corpus and Cross-Architecture Analysis}
\label{sec:cross_corpus}

\subsection{Distributional Landscape}
\label{ssec:landscape}

\begin{figure}[t]
    \centering
    \includegraphics[width=0.88\linewidth]{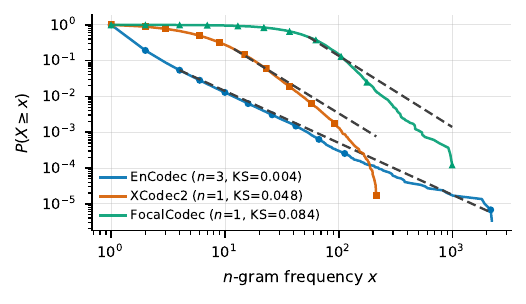}
    \caption{Empirical complementary CDFs of the $n$-gram type frequencies (solid, with markers) and fitted power laws above $x_{\min}$ (dashed) for one representative codec per meta-category on clean LJSpeech at the 500K-token reference; larger KS corresponds to larger visible departure from the fitted line.}
    \label{fig:dist_fit_examples}
\end{figure}

\begin{figure}[t]
    \centering
    \includegraphics[width=0.95\linewidth]{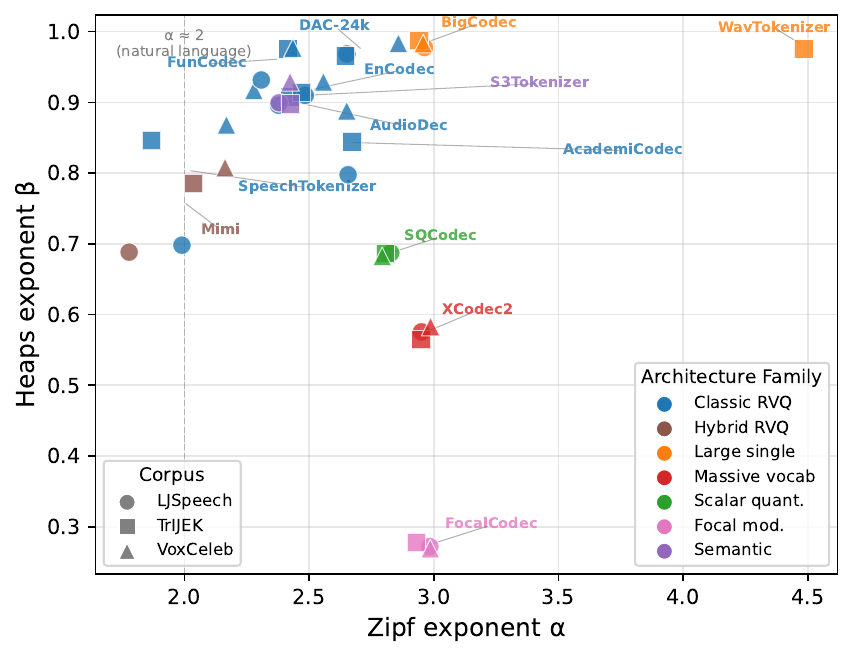}
    \caption{Clean-speech Zipf exponent $\alpha$ and Heaps exponent $\beta$ for 13 codecs on LJSpeech, VoxCeleb, and TrIJEK at the 500K-token reference scale. Colors denote the architecture families of Table~\ref{tab:codecs}. Cells without a defined power-law fit ($V/N>0.9$) are omitted.}
    \label{fig:multicorpus}
\end{figure}

Figure~\ref{fig:multicorpus} shows the clean-speech $\alpha$--$\beta$ landscape of 13 codecs on LJSpeech, VoxCeleb, and TrIJEK at the 500K-token reference scale.
The estimates use the codec-specific $n$-gram orders adopted in Section~\ref{ssec:ngram_ablation}.
Corpus-level variation is limited within the multi-codebook RVQ family.
The RVQ-family averages are similar for LJSpeech ($\alpha=2.32$, $\beta=0.84$), VoxCeleb ($\alpha=2.44$, $\beta=0.91$), and TrIJEK ($\alpha=2.42$, $\beta=0.91$).
Table~\ref{tab:dist_params} lists the per-codec parameters underlying this landscape, averaged over the corpora with valid fits. Figure~\ref{fig:dist_fit_examples} shows the empirical frequency distributions and fitted power laws behind three representative rows, visualizing what small and large KS distances correspond to.

\begin{table}[t]
\caption{Distributional parameters of the 13 codecs at the 500K-token reference: the Zipf exponent $\alpha$ with its KS fit distance, and the Heaps exponent $\beta$ with its intercept $K$. Values are averaged over the $N_{\mathrm{fit}}$ corpora with valid fits, use the codec-specific analysis orders under the fitting safeguards of Section~\ref{ssec:metrics}, and are reported descriptively.}
\label{tab:dist_params}
\centering
\scriptsize
\begin{tabular}{@{}llccccc@{}}
\toprule
\textbf{Codec} & \textbf{Meta-cat.} & $N_{\mathrm{fit}}$ & $\alpha$ & KS & $\beta$ & $K$ \\
\midrule
Mimi & RVQ & 2 & 1.97 & 0.021 & 0.75 & 7.8 \\
SpeechTokenizer & RVQ & 3 & 2.01 & 0.014 & 0.80 & 4.0 \\
AudioDec & RVQ & 3 & 2.36 & 0.005 & 0.91 & 2.0 \\
EnCodec & RVQ & 3 & 2.50 & 0.005 & 0.92 & 1.9 \\
FunCodec & RVQ & 3 & 2.39 & 0.011 & 0.96 & 1.3 \\
AcademiCodec & RVQ & 3 & 2.66 & 0.009 & 0.84 & 3.8 \\
DAC-24k & RVQ & 3 & 2.72 & 0.010 & 0.97 & 1.3 \\
\midrule
XCodec2 & Single VQ & 3 & 2.96 & 0.050 & 0.57 & 41.7 \\
BigCodec & Single VQ & 2 & 2.88 & 0.179 & 0.98 & 1.2 \\
WavTokenizer & Single VQ & 1 & 4.48 & 0.002 & 0.97 & 1.3 \\
\midrule
FocalCodec & Non-VQ & 3 & 2.97 & 0.093 & 0.27 & 286.9 \\
SQCodec & Non-VQ & 3 & 2.81 & 0.083 & 0.69 & 18.5 \\
S3Tokenizer & Non-VQ & 3 & 2.41 & 0.004 & 0.91 & 2.0 \\
\bottomrule
\end{tabular}
\end{table}

\textbf{Architecture families occupy distinct regions of the distributional space.}
Under the adopted analysis orders, multi-codebook RVQ codecs span $\alpha\approx2.0$--$2.7$, with the semantically informed Mimi and SpeechTokenizer at the low end (Table~\ref{tab:dist_params}).
Among the non-VQ codecs, SQCodec and FocalCodec sit near $\alpha\approx2.8$--$3.0$ under their adopted orders, adjacent to XCodec2 at its unigram order.
The semantic tokenizer S3Tokenizer is lower at 2.41.
FocalCodec has the lowest Heaps exponent in the codec set, reflecting its near-complete use of the 8{,}192-entry codebook within each 500K-token sample.
Among valid clean-speech fits, $\alpha$ is lowest for Mimi and highest for WavTokenizer.

\subsection{Entropy, Redundancy, and Perplexity}
\label{ssec:entropy}

Table~\ref{tab:entropy} reports the unigram Shannon entropy, redundancy, and token perplexity of Section~\ref{ssec:metrics}, computed on the deduplicated token streams at the 500K-token reference.
The common unigram order keeps the values comparable across codecs.

\textbf{On clean speech, entropy tracks vocabulary scale and redundancy is uniformly low.}
Entropy is positively associated with the available vocabulary scale in this codec set.
Every codec uses its observed inventory close to uniformly, with the single-codebook VQ group and FocalCodec lowest.
These values sit far below classical redundancy estimates for natural-language text ($R\approx0.5$ for printed English), which include the sequential dependencies that a unigram measure ignores~\cite{shannon1951}.
The token perplexity $2^H$ varies by three orders of magnitude across the codec set at a matched data scale.

\textbf{White noise removes entropy; DEMAND noise barely changes it.}
About 1--4.4 bits vanish under white noise for every codec except S3Tokenizer and SQCodec, whose entropy changes by less than 0.5 bits.
Probability mass concentrates on a smaller effective inventory, raising the grid-mean redundancy from 0.08 on clean speech to 0.27.
DEMAND noise changes entropy comparatively little.

\begin{table}[t]
\caption{Unigram Shannon entropy $H$ (bits), redundancy $R$, and token perplexity $2^H$ on clean speech, with the change in $H$ from clean speech to each 0\,dB condition. Values are computed on deduplicated streams at the 500K-token reference and averaged over the corpora with a matched reference (the Mimi--TrIJEK cells are excluded).}
\label{tab:entropy}
\centering
\scriptsize
\begin{tabular}{@{}lccccc@{}}
\toprule
\textbf{Codec} & $H$ & $R$ & $2^H$ & $\Delta H_{\text{White}}$ & $\Delta H_{\text{DEMAND}}$ \\
\midrule
Mimi            & 10.2 & 0.058 & 1{,}232  & $-2.67$ & $-0.31$ \\
SpeechTokenizer & 9.3  & 0.063 & 654      & $-4.14$ & $-1.54$ \\
AudioDec        & 7.4  & 0.237 & 172      & $-4.35$ & $-0.53$ \\
EnCodec         & 8.0  & 0.162 & 250      & $-2.90$ & $+0.26$ \\
FunCodec        & 9.3  & 0.073 & 619      & $-4.40$ & $-0.66$ \\
AcademiCodec    & 6.3  & 0.120 & 82       & $-0.99$ & $-0.26$ \\
DAC-24k         & 8.9  & 0.108 & 486      & $-2.12$ & $+0.45$ \\
\midrule
XCodec2         & 15.2 & 0.043 & 36{,}830 & $-3.33$ & $-0.87$ \\
BigCodec        & 12.7 & 0.023 & 6{,}648  & $-3.48$ & $-0.61$ \\
WavTokenizer    & 11.5 & 0.041 & 2{,}883  & $-3.47$ & $-1.04$ \\
\midrule
FocalCodec      & 12.6 & 0.029 & 6{,}309  & $-1.09$ & $-0.19$ \\
SQCodec         & 16.3 & 0.027 & 83{,}077 & $-0.08$ & $+0.05$ \\
S3Tokenizer     & 9.3  & 0.049 & 634      & $-0.47$ & $+0.15$ \\
\bottomrule
\end{tabular}
\end{table}

\subsection{Clean-to-Noisy Distributional Shifts}
\label{ssec:noise_robustness_results}

\begin{figure*}[t]
    \centering
    \includegraphics[width=0.98\textwidth]{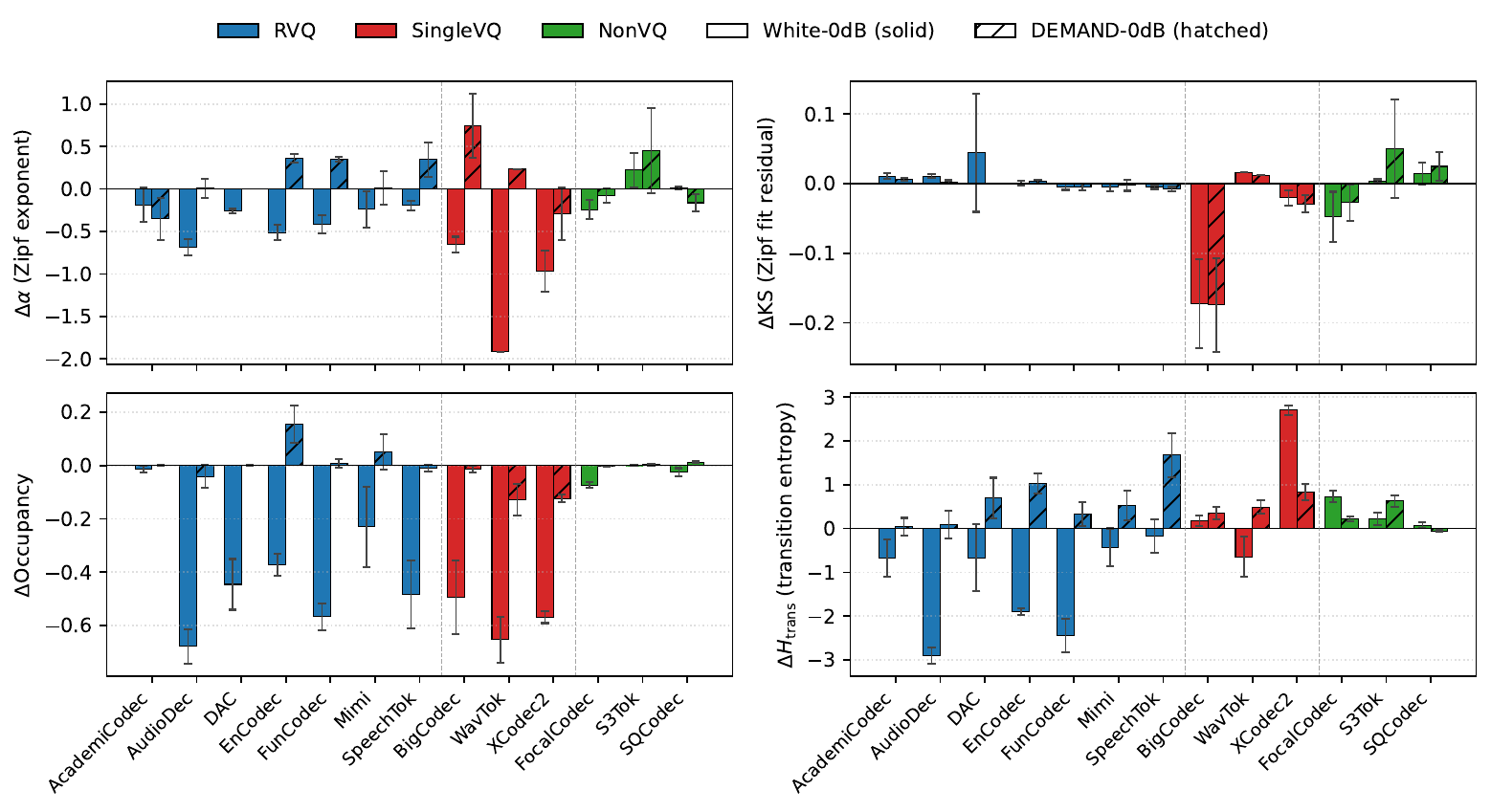}
    \caption{Per-codec clean-to-noise shifts in the Zipf exponent, KS distance, codebook occupancy, and transition entropy (Section~\ref{ssec:mechanism_axes}) under the two 0\,dB conditions (solid: white; hatched: DEMAND). Bars show means across the three corpora and error bars the standard deviation; colors denote the quantizer meta-category, and dashed vertical lines separate the meta-categories.}
    \label{fig:clean_to_0db_structural}
\end{figure*}

For each codec, corpus, and distributional metric $m$, the $\alpha$, KS, $\beta$, $K$, $H$, and $R$ of Section~\ref{ssec:metrics}, together with the sequence diagnostics of Section~\ref{ssec:mechanism_axes}, we calculate the shift from clean speech to 0\,dB noise as $\Delta m=m_{\mathrm{0dB}}-m_{\mathrm{clean}}$.
Figure~\ref{fig:clean_to_0db_structural} reports the codec-level shifts averaged over the three corpora.
The median absolute clean-to-0\,dB change in $\alpha$ is 0.35 under white noise and 0.23 under DEMAND noise, with the most affected cells shifting by up to 1.91.

\textbf{White noise depresses $\alpha$ most strongly for the single-codebook VQ codecs.}
Among the six single-codebook cells with valid clean and white-noise fits, the group mean is $\overline{\Delta\alpha}=-1.02$, compared with $-0.36$ for the multi-codebook RVQ cells (20 cells) and $0.00$ for the non-VQ cells (nine cells).
DEMAND-noise shifts are smaller on average and vary more across codecs and corpora than white-noise shifts.

\textbf{No single noise-robustness ranking of codecs exists: cross-noise agreement is metric-dependent.}
To test whether the two noise types stress the same codecs in the same order, we compare, for each metric, the codec rankings induced under white and under DEMAND noise.
UER-based codec rankings are similar across the two noise types (Spearman $\rho=0.81$, $p=0.001$ across the 13 codecs).
Occupancy-shift rankings show moderate agreement ($\rho=0.65$, $p=0.015$, 13 codecs), while $\alpha$-shift rankings show no agreement ($\rho=-0.04$, $p=0.90$, the 12 codecs with valid fits).
Therefore, robustness rankings are metric-dependent: token-identity disruption (UER) is consistent across noise types, whereas distribution-shape sensitivity ($\Delta\alpha$) is noise-type-specific.


\subsection{Variance Decomposition}
\label{ssec:anova}

We fit a full-factorial three-way ANOVA using one 100K-token aggregate per valid codec--corpus--condition cell of the 117-cell grid.
Because validity filtering yields an unbalanced design, effect sizes are computed from Type-II sums of squares.
Table~\ref{tab:anova} reports eta-squared values, where $\eta^2$ is the fraction of total variance associated with each term, together with the number of valid cells per metric.

\textbf{The dominant factor differs across metrics.}
Acoustic condition is the largest main effect for $\alpha$, $\beta$, and the redundancy $R$.
Meta-category leads for KS, the Heaps intercept $K$, and most strongly for the unigram entropy $H$, which is the most architecture-determined metric in the panel.
Corpus identity explains little variation in the evaluated three-corpus grid.

The meta$\times$condition interaction accounts for an appreciable variance fraction for $\alpha$, KS, and $R$, indicating that the effect of acoustic condition differs across quantizer meta-categories; Section~\ref{sec:mechanism} characterizes these family-specific responses as sequence-level degradation signatures.
A large residual component remains for all metrics.
The reported values therefore describe variation associated with the grouped factors; differences among individual codecs within a meta-category are absorbed by the residual.

\textbf{The noise effects survive a codec random intercept.}
Because the cells repeatedly sample the same 13 codecs, we refit each metric with a linear mixed model comprising the meta$\times$condition and corpus fixed effects and a codec random intercept, which absorbs a large variance share (intraclass correlation 0.46--0.96).
Under this model, the joint noise effect---condition together with its meta-category interaction---remains significant for every metric (Wald $p\leq0.001$), as does the meta$\times$condition interaction alone ($p\leq0.001$).
The meta-category contrast evaluated on clean speech remains significant for $\alpha$, $H$, and $R$.
The corpus terms reach significance only for $\beta$, with a test statistic well below the noise terms.

\begin{table}[t]
\caption{Type-II eta-squared values from the full-factorial three-way ANOVA. $N_{\mathrm{cells}}$ is the number of valid cells per metric. MC denotes the Meta-Codec quantizer category, Crp.\ the corpus, and Cnd.\ the acoustic condition; the tabulated terms partition the total variance, so each row sums to one.}
\label{tab:anova}
\centering
\scriptsize
\setlength{\tabcolsep}{2.5pt}
\begin{tabular}{@{}lccccccccc@{}}
\toprule
\textbf{Metric} & \textbf{$N_{\mathrm{cells}}$} & \textbf{MC} & \textbf{Crp.} & \textbf{Cnd.} & \makecell{\textbf{MC}\\$\times$\textbf{Cnd.}} & \makecell{\textbf{MC}\\$\times$\textbf{Crp.}} & \makecell{\textbf{Crp.}\\$\times$\textbf{Cnd.}} & \makecell{\textbf{MC}$\times$\textbf{Crp.}\\$\times$\textbf{Cnd.}} & \textbf{Resid.} \\
\midrule
$\alpha$ & 100 & 0.134 & 0.003 & 0.278 & 0.088 & 0.020 & 0.015 & 0.014 & 0.449 \\
KS & 100 & 0.158 & 0.003 & 0.016 & 0.096 & 0.012 & 0.002 & 0.009 & 0.704 \\
$\beta$ & 117 & 0.086 & 0.005 & 0.120 & 0.039 & 0.002 & 0.002 & 0.001 & 0.745 \\
$K$ & 117 & 0.240 & 0.001 & 0.000 & 0.016 & 0.000 & 0.001 & 0.000 & 0.742 \\
$H$ & 117 & 0.551 & 0.000 & 0.126 & 0.023 & 0.000 & 0.001 & 0.000 & 0.299 \\
$R$ & 117 & 0.320 & 0.002 & 0.351 & 0.080 & 0.001 & 0.004 & 0.001 & 0.241 \\
\bottomrule
\end{tabular}
\end{table}

\section{Distributional Structure and Quality}
\label{sec:quality}

\subsection{Clean-Speech Quality Overview}
\label{ssec:codec_quality}

Table~\ref{tab:codec_quality} summarizes the quality metrics of the 12 decodable codecs, averaged over the three clean corpora.
S3Tokenizer does not provide reconstructed audio, so its quality metrics are unavailable.

\begin{table}[t]
\caption{Clean-speech resynthesis quality for the 12 decodable codecs, averaged over the three clean corpora. The first row gives the UTMOS of the reference recordings themselves; their reference-based metrics are zero by definition. S3Tokenizer is excluded because it provides no decoder; UER is omitted because it is defined only under noisy conditions.}
\label{tab:codec_quality}
\centering
\scriptsize
\begin{tabular}{@{}lccccc@{}}
\toprule
\textbf{Codec} & \textbf{UTMOS}$\uparrow$ & \makecell{\textbf{MCD}$\downarrow$\\(dB)} & \makecell{\textbf{F0RMSE}$\downarrow$\\(Hz)} & \makecell{\textbf{F0MedAE}$\downarrow$\\(Hz)} & \textbf{dCER}$\downarrow$ \\
\midrule
\textit{Reference} & \textit{3.09} & --- & --- & --- & --- \\
\midrule
Mimi & 3.05 & 3.1 & 15.7 & 0.8 & 0.04 \\
SpeechTokenizer & 2.96 & 3.9 & 21.9 & 1.3 & 0.04 \\
AudioDec & 2.34 & 4.0 & 25.2 & 2.2 & 0.04 \\
EnCodec & 2.26 & 3.2 & 19.0 & 0.8 & 0.02 \\
FunCodec & 2.93 & 3.0 & 24.3 & 1.1 & 0.02 \\
AcademiCodec & 2.94 & 3.7 & 24.7 & 1.0 & 0.04 \\
DAC-24k & 3.06 & 1.3 & 7.0 & 0.1 & 0.01 \\
\midrule
XCodec2 & 3.32 & 3.9 & 23.8 & 1.2 & 0.04 \\
BigCodec & 3.32 & 3.6 & 23.2 & 1.0 & 0.06 \\
WavTokenizer & 2.74 & 5.0 & 24.3 & 1.5 & 0.21 \\
\midrule
FocalCodec & 3.33 & 5.1 & 28.3 & 3.6 & 0.05 \\
SQCodec & 3.08 & 3.5 & 21.0 & 0.6 & 0.02 \\
\bottomrule
\end{tabular}
\end{table}

\textbf{Clean-speech quality sharply separates the codec set.}
The clean-speech dCER draws the sharpest line: every codec except WavTokenizer preserves intelligibility with dCER of at most 0.06, whereas WavTokenizer reaches 0.21.
This degradation appears at the transcription level rather than in the frame-level acoustic distances: its MCD and F0 errors stay within the range of the remaining codecs.

\begin{figure}[t]
    \centering
    \includegraphics[width=0.98\linewidth]{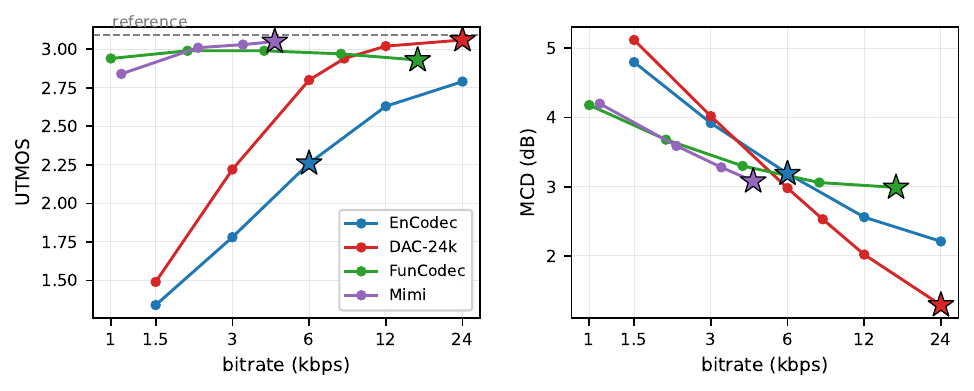}
    \caption{Clean-speech UTMOS and MCD versus bitrate for the four RVQ codecs whose bandwidth is an inference-time choice. Stars mark the configurations evaluated in this paper, and the dashed line gives the UTMOS of the reference.}
    \label{fig:bitrate_sweep}
\end{figure}

\textbf{Faithful decoders are bounded by their reference; regenerative decoders exceed it.}
The remaining spread in Table~\ref{tab:codec_quality} follows from bitrate and decoder design.
For example, DAC-24k's resynthesis is close to transparent in terms of MCD and F0 error, while its per-corpus UTMOS matches that of the reference recordings to within 0.07.
Therefore, its reference-fidelity metrics are the best in the table, while its UTMOS is bounded by the naturalness of the references themselves.
XCodec2 and BigCodec instead exceed the reference UTMOS on TrIJEK and VoxCeleb by up to 0.5. This shows that their decoders regenerate speech toward the clean characteristics of their training data, thereby raising naturalness while increasing the reference-based spectral distance.
Figure~\ref{fig:bitrate_sweep} places the adopted configurations of the four RVQ codecs with an inference-time bandwidth control on their clean-speech quality--bitrate curves.
The first-level token streams are identical across these configurations, so all distributional analyses are unaffected by this choice.

\vspace{-2mm}

\subsection{JSD and Quality under Noise}
\label{ssec:jsd_comparison}

We measure distributional degradation using the JSD between equal-sized clean and noisy token samples of each codec--corpus cell, with the sample size fixed to the largest value available for every evaluated cell ($\approx$423K tokens).
JSD is computed in token space and requires a clean reference distribution, but it does not require waveform reconstruction.
Because JSD approaches its upper bound when the two samples share little $n$-gram support, we compute it at the common unigram order ($n=1$) for every codec---deliberately departing from the codec-specific analysis orders of Section~\ref{ssec:ngram_ablation}: under those orders JSD saturates above $0.9$ for most higher-order codecs and leaves little cross-codec variation, and the unigram order is the only order valid across the full grid.

\begin{figure}[t]
    \centering
    \includegraphics[width=0.98\linewidth]{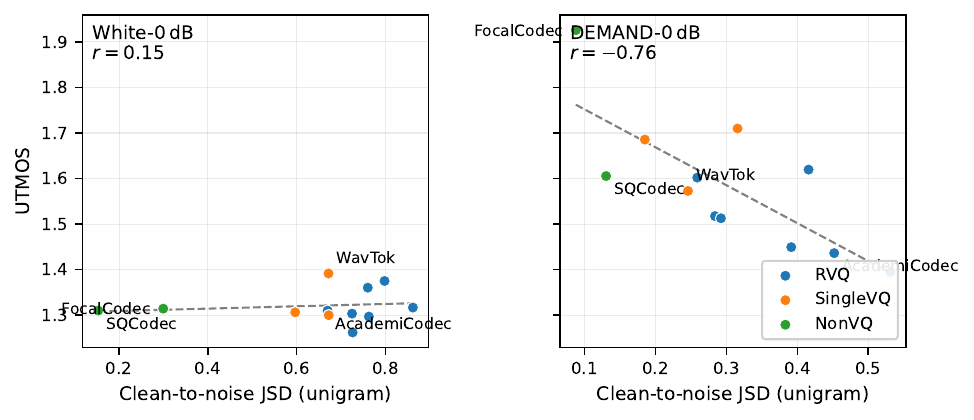}
    \caption{Clean-to-noise JSD at the common unigram order versus UTMOS under the corresponding noisy condition, for the twelve decodable codecs averaged across the three corpora. Dashed lines are least-squares fits; $r$ is the Pearson correlation reported in Table~\ref{tab:correlations}.}
    \label{fig:jsd_utmos}
\end{figure}

Table~\ref{tab:correlations} reports Pearson and Spearman correlations between JSD and the quality metrics measured under the corresponding noisy condition, using the twelve decodable codecs after averaging each codec--noise cell across the three corpora.
Figure~\ref{fig:jsd_utmos} shows the corresponding JSD--UTMOS scatter under the two noisy conditions.

\textbf{JSD tracks perceived quality most clearly under DEMAND noise.}
The association between JSD and quality depends on noise type and on the statistic used.
Under DEMAND noise, cells with a larger distributional shift have lower UTMOS and a larger median F0 error, whereas the MCD association is weak once the level correction of Section~\ref{ssec:quality_metrics} is applied.
The UTMOS association survives removing the two non-VQ codecs that occupy the low-JSD end ($r=-0.68$, $p=0.029$), and in the scatter of Figure~\ref{fig:jsd_utmos} no single codec moves the coefficient by more than 0.10.
Under white noise no association with the audio-based quality metrics survives both statistics: the MCD coefficient holds only in Pearson terms and reverses in sign ($r=-0.27$) without the two non-VQ codecs, reflecting the gap between those codecs and the rest rather than a graded relation.
The dCER associations are weak throughout, and the negative UER association, although present under white noise in both statistics, vanishes under DEMAND noise.

\textbf{Distribution-level shift and token-level integrity capture different aspects of degradation.}
The UER cell means span 0.66--1.00, so the weak JSD--UER association under DEMAND noise is partly attributable to the uniformly high token-level degradation at 0\,dB.

\begin{table}[t]
\caption{Pearson ($r$) and Spearman ($\rho$) correlations between clean-to-noise JSD (common unigram order, equal-sized samples) and quality metrics under the corresponding noisy condition. Cells are the twelve decodable codecs averaged across the three corpora (12 codecs per condition). Bold values indicate a magnitude above 0.5.}
\label{tab:correlations}
\centering
\scriptsize
\begin{tabular}{@{}llccccc@{}}
\toprule
\textbf{Condition} & \textbf{Stat.} & \textbf{UTMOS} & \textbf{MCD} & \textbf{F0MedAE} & \textbf{UER} & \textbf{dCER} \\
\midrule
White-0\,dB & $r$ & $\phantom{-}0.15$ & $\mathbf{\phantom{-}0.62}$ & $\phantom{-}0.19$ & $\mathbf{-0.56}$ & $-0.01$ \\
 & $\rho$ & $\phantom{-}0.06$ & $\phantom{-}0.06$ & $-0.17$ & $\mathbf{-0.73}$ & $\mathbf{-0.50}$ \\
\midrule
DEMAND-0\,dB & $r$ & $\mathbf{-0.76}$ & $\phantom{-}0.38$ & $\mathbf{\phantom{-}0.57}$ & $-0.03$ & $-0.14$ \\
 & $\rho$ & $\mathbf{-0.65}$ & $\phantom{-}0.49$ & $\phantom{-}0.34$ & $-0.23$ & $-0.25$ \\
\bottomrule
\end{tabular}
\end{table}

\section{Architecture-Dependent Degradation Signatures}
\label{sec:mechanism}

We analyze whether noise produces different sequence-level degradation patterns across quantizer architectures.

\vspace{-2mm}

\subsection{Operational Definitions}
\label{ssec:mechanism_axes}

We characterize each clean-to-noise shift using two metrics: repetition rate and transition entropy.
We define repetition rate $r$, which is the fraction of adjacent positions that contain the same token before deduplication.
We also define transition entropy $H_{\text{trans}}=H(t_i\mid t_{i-1})$, which measures the uncertainty of the next token under the empirical bigram distribution.

For each codec, corpus, and noise type, we calculate $\Delta r$ and $\Delta H_{\text{trans}}$ between clean speech and the corresponding 0\,dB condition.
We define collapse as $\Delta r>0.01$ and $\Delta H_{\text{trans}}<-0.1$, and explosion as $\Delta r<-0.01$ and $\Delta H_{\text{trans}}>0.1$.
All remaining cells are classified as neutral. These deadband thresholds are operational choices that exclude near-zero shifts.
Both diagnostics are computed on the original streams before consecutive deduplication.

\vspace{-2mm}

\subsection{Shifts by Quantizer Architecture}
\label{ssec:mechanism_shift}

\begin{figure}[t]
    \centering
    \includegraphics[width=0.98\linewidth]{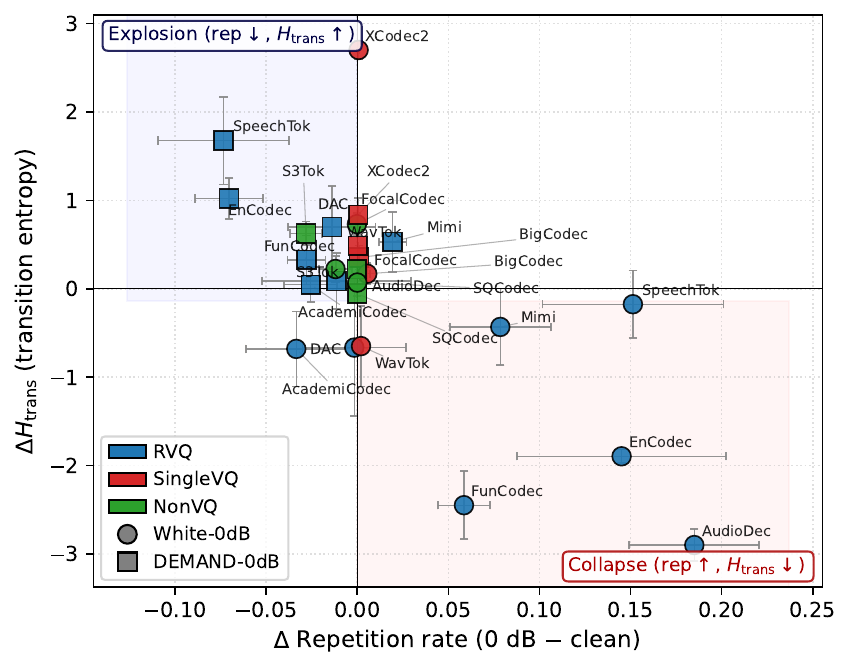}
    \caption{Per-codec shifts from clean speech to the 0\,dB conditions in repetition rate and transition entropy, averaged across the three corpora. Colors denote quantizer meta-category and markers the noise type; shaded quadrants indicate the collapse and explosion signatures.}
    \label{fig:clean_to_0db_mechanism}
\end{figure}

Figure~\ref{fig:clean_to_0db_mechanism} complements the view of Figure~\ref{fig:clean_to_0db_structural} by placing each codec--noise pair on the signature plane spanned by the repetition-rate and transition-entropy shifts. Table~\ref{tab:mechanism_class} summarizes the group-mean shifts.

\textbf{White noise induces collapse, and DEMAND noise explosion, within the same RVQ family.}
White noise produces a collapse-like pattern in the multi-codebook RVQ group: repetition rises while transition entropy and codebook occupancy drop sharply, mirroring the unigram entropy loss reported in Section~\ref{ssec:entropy}.
The same group responds to DEMAND noise in the opposite direction, with decreasing repetition, increasing transition entropy, and little average change in occupancy.
In the signature plane of Fig.~\ref{fig:clean_to_0db_mechanism}, the RVQ--white points accordingly cluster in the collapse quadrant, and the RVQ--DEMAND points in the explosion quadrant.

\textbf{Single-codebook VQ codecs deform without either signature.}
They show little change in repetition rate under either noise type, remaining near the $\Delta r=0$ axis of the signature plane, while losing 33--75 percentage points of occupancy per cell under white noise.
These changes do not satisfy the collapse or explosion criteria.
Their observed shifts appear in occupancy and distribution-shape parameters rather than in repetition rate.

The non-VQ group does not show a collapse signature under either noise type.
Transition entropy generally increases while repetition-rate changes remain small, but the magnitude differs sharply within the group.
S3Tokenizer crosses the explosion deadband in four of its six cells, whereas FocalCodec and SQCodec change too little to leave the neutral region.

\vspace{-2mm}

\subsection{Distribution of Degradation Signatures}
\label{ssec:mechanism_distribution}

\textbf{Collapse concentrates in RVQ--white and explosion in RVQ--DEMAND, robustly to the deadband choice.}
Table~\ref{tab:mechanism_class} reports the classification of the 78 codec--corpus--noise cells under the deadband criteria of Section~\ref{ssec:mechanism_axes}.
Collapse is concentrated in the RVQ--white condition, which accounts for 14 of the 15 collapse instances in the full analysis.
Explosion is concentrated in the RVQ--DEMAND condition, consistent with the group-average increase in transition entropy under DEMAND noise.
The single RVQ--white explosion cell belongs to DAC-24k on TrIJEK.
Halving both thresholds changes the RVQ--white collapse and RVQ--DEMAND explosion counts to 15 and 15 of the 21 cells in each group, while doubling them yields 14 and 11.

All 18 SingleVQ cells are classified as neutral.
The non-VQ group contributes a further four explosion cells, all belonging to S3Tokenizer.
The semantic tokenizer satisfies the explosion criterion in all three corpora under DEMAND noise and in one under white noise, whereas FocalCodec and SQCodec are neutral throughout.

\begin{table}[t]
\caption{Degradation-signature classification of the 78 codec--corpus--noise cells under the deadband criteria of Section~\ref{ssec:mechanism_axes}. \textbf{Col.}/\textbf{Exp.}/\textbf{Neu.}: number of cells classified as collapse, explosion, and neutral; $\overline{\Delta r}$, $\overline{\Delta H_{\text{trans}}}$, $\overline{\Delta\text{Occ.}}$: group-mean clean-to-noise shifts in repetition rate, transition entropy, and codebook occupancy (percentage points).}
\label{tab:mechanism_class}
\centering
\scriptsize
\begin{tabular}{@{}llcccccc@{}}
\toprule
\textbf{Meta-cat.} & \textbf{Noise} & \textbf{Col.} & \textbf{Exp.} & \textbf{Neu.} & $\overline{\Delta r}$ & $\overline{\Delta H_{\text{trans}}}$ & $\overline{\Delta\text{Occ.}}$ \\
\midrule
RVQ & White & 14 & 1 & 6 & $+0.08$ & $-1.31$ & $-40$ \\
RVQ & DEMAND & 1 & 13 & 7 & $-0.03$ & $+0.63$ & $+2$ \\
SingleVQ & White & 0 & 0 & 9 & $+0.00$ & $+0.74$ & $-57$ \\
SingleVQ & DEMAND & 0 & 0 & 9 & $+0.00$ & $+0.56$ & $-9$ \\
NonVQ & White & 0 & 1 & 8 & $-0.00$ & $+0.34$ & $-3$ \\
NonVQ & DEMAND & 0 & 3 & 6 & $-0.01$ & $+0.27$ & $+1$ \\
\midrule
\textbf{Total} & --- & \textbf{15} & \textbf{18} & \textbf{45} & --- & --- & --- \\
\bottomrule
\end{tabular}
\end{table}

\section{Discussion}
\label{sec:discussion}

\subsection{Architecture and Condition as the Dominant Distributional Factors}
\label{ssec:disc_arch}

A consistent thread across our results is that quantizer architecture exerts a far larger influence on token-level statistics than any of the conventional sources of corpus variation we tested, with corpus identity contributing negligibly across all distributional parameters (Section~\ref{ssec:anova}).
The Zipf exponent $\alpha$ spans from approximately 2.0 to above 4 on clean speech across the 13-codec set (Section~\ref{ssec:landscape}), a range larger than the typical clean$\rightarrow$0\,dB shift of individual codecs.
At the same time, the three-way ANOVA (Section~\ref{ssec:anova}) shows that architecture does not dominate uniformly.
Meta-category is the leading main effect for the fit quality KS, the Heaps intercept $K$, and most strongly for the unigram entropy $H$.
Acoustic condition dominates $\alpha$, $\beta$, and the redundancy $R$ at the pooled level.
The meta$\times$condition interaction remains significant for every metric once codec identity is modeled as a random effect.
This pattern motivates a methodological shift relative to~\cite{park_asru2025}, where a single corpus was sufficient to characterize a codec.
In our expanded grid, any cross-codec comparison must be conducted at fixed meta-category, $n$-gram order, and where possible vocabulary scale, because between-family contrasts are confounded with all three.

A practical implication is that the natural-language analogy that motivated the original Zipf analysis of NAC tokens---namely, that an $\alpha$ near the word-level reference value of 2 reflects language-like statistical structure---should be interpreted family-conditionally rather than as a universal target.
Throughout this paper, \emph{language-like} refers to the qualitative statistical structure that codec tokens share with text---a well-fitting power-law frequency distribution and sublinear Heaps-type vocabulary growth that persist across corpora---not to quantitative agreement with word-level parameter values, which the differences in unit granularity and vocabulary size preclude.
Under this reading, a codec is language-like to the extent that its fits satisfy the validity and goodness-of-fit criteria of Section~\ref{ssec:metrics} across corpora, and every codec in the panel meets this criterion on clean speech. 


\vspace{-2mm}

\subsection{JSD as a Complementary Distributional Diagnostic}
\label{ssec:disc_jsd}

Among the distributional diagnostics we considered, JSD shows the clearest association with acoustic quality under real-life DEMAND noise, where it tracks UTMOS and the median F0 error in the expected directions (Section~\ref{ssec:jsd_comparison}).
JSD operates on the full token frequency distribution, so probability-mass redistribution that preserves the overall power-law shape but alters individual token frequencies---a common consequence of noise---registers in JSD but not in $\alpha$ or KS.
Because JSD is computed from token-level counts and requires no waveform reconstruction, it can also be evaluated for non-decodable tokenizers such as S3Tokenizer, for which audio-based quality metrics are unavailable.

\vspace{-2mm}

\subsection{Codebook Size and the Distributional Regime}
\label{ssec:disc_vocab}

The multi-codebook RVQ codecs, whose per-level codebooks hold 1{,}024 or 2{,}048 entries, occupy the low-$\alpha$ end of the spectrum.
Every waveform-reconstructing design built on a single codebook of 4{,}096 entries or more sits above 2.8, with WavTokenizer highest at 4.48.
The semantic tokenizer S3Tokenizer, whose codebook is trained for recognition rather than reconstruction, sits lower at 2.41 despite its 4{,}096 entries.
This ordering is consistent with a sampling argument: as $|V|$ grows at a fixed corpus size, the frequency profile is increasingly dominated by a small set of acoustically frequent units, with a long tail of rarely used codewords that steepens the fitted exponent.
The degradation-signature analysis (Section~\ref{sec:mechanism}) is consistent with the same axis: single-codebook VQ codecs respond to noise predominantly via probability-mass redistribution within the codebook (large $\Delta$Occupancy, small $\Delta r$), while multi-codebook RVQ codecs respond via local sequence dynamics (large $\Delta r$, large $\Delta H_{\text{trans}}$).

\vspace{-2mm}

\subsection{Limitations}
\label{ssec:disc_limitations}

Several limitations should be considered when generalizing our findings.
First, cross-codec comparisons are mediated by family-conditional $n$-gram orders, so absolute parameter values are not directly comparable across families (Section~\ref{ssec:ngram_ablation}).
Second, the noise design is restricted to two source categories at a single 0\,dB SNR, leaving other SNRs, reverberation, and channel distortions unexplored.
Third, the quality metrics rely on automatic estimators (Whisper-large-v3, UTMOS) whose biases may correlate with codec architecture.
Fourth, S3Tokenizer lacks a decoder, so the JSD--quality analyses of Section~\ref{ssec:jsd_comparison} cover 12 of the 13 codecs.
Fifth, the ANOVA cells repeatedly sample the same 13 codecs, so the reported $\eta^2$ values are descriptive partitions and the mixed-model meta-category tests rest on 13 codec-level units.
Sixth, the JSD--quality correlations are computed over twelve codec-level means, a sample size at which the choice of statistic matters (Section~\ref{ssec:jsd_comparison}).
Seventh, cross-codec JSD requires a matched low $n$-gram order because the measure saturates when the compared samples share little $n$-gram support.
Eighth, vocabulary size is confounded with architecture, bitrate, and training objective, so the codebook-size patterns of Section~\ref{ssec:disc_vocab} are associations rather than isolated effects.
Ninth, the collapse and explosion counts depend on the operational deadband thresholds, within the two-fold sensitivity range reported in Section~\ref{ssec:mechanism_distribution}.
Finally, we treat each codec as a fixed black-box pipeline, leaving the relation between training objective or data composition and the measured distributional regime to future work.

\section{Conclusion}
\label{sec:conclusion}

We presented a language-statistical analysis of 13 neural audio codecs spanning multi-codebook RVQ, single-codebook VQ, and non-VQ designs, evaluated on three corpora and three acoustic conditions in a fully crossed 117-cell grid.

The analysis supports three principal conclusions.
First, corpus identity is a negligible source of distributional variance: acoustic condition and quantizer architecture dominate in a metric-dependent way.
The significant meta--condition interaction indicates family-specific noise responses.
Second, JSD computed at a common unigram order from equal-sized token samples provides a decoder-free distribution-level diagnostic.
It tracks perceived quality across codecs under real-world DEMAND noise, whereas none of its white-noise associations survives a rank-based check.
Third, the collapse and explosion degradation signatures previously identified for RVQ codecs do not generalize uniformly.
Collapse concentrates in RVQ$\times$white-noise cells, and explosion in RVQ$\times$DEMAND cells and the semantic tokenizer.
Single-codebook VQ codecs deform along occupancy and $\alpha$ without either signature.

Beyond these findings, the paper contributes an architecture-conditioned protocol for applying language-statistical analysis to codec tokens: family-conditional $n$-gram orders with explicit fit-validity safeguards, matched-sample chunk-based reliability assessment, and variance decomposition across meta-category, corpus, and condition.
This protocol makes the natural-language analogy underlying prior Zipf analyses of codec tokens usable as a family-conditional convention rather than a universal target.

Future work should broaden the noise design beyond the 0\,dB clean$\rightarrow$noisy contrast, to continuous SNR sweeps and to non-stationary distortions including reverberation and channel effects.
The signature analysis should also extend to emerging codec families, in particular large-vocabulary and semantic-prior designs whose distributional behavior under noise remains underexplored.
A further direction is to connect the distributional regimes characterized here to the training objective and data composition of each codec.
We hope that the metric panel, reliability protocol, and analysis conventions established provide a useful template for such studies.

\textbf{Acknowledgment:}
The work was supported by JSPS KAKENHI Grant Number 26KJ0771, JST Moonshot Grant Number JPMJMS2011,  and based on results from a project outsourced by the New Energy and Industrial Technology Development Organization (NEDO), and the BRIDGE Program (R7-H05), implemented by the Cabinet Office, Government of Japan.
\vspace{-2mm}

\bibliographystyle{IEEEtran}
\bibliography{references}


\end{document}